\documentclass[letterpaper, 10 pt, conference]{ieeeconf}  

\IEEEoverridecommandlockouts                              

\usepackage{cite}
\usepackage{amsmath,amssymb,amsfonts}
\usepackage{algorithmic}
\usepackage{algorithm}
\usepackage{graphicx}
\usepackage{textcomp}
\usepackage{subcaption}
\usepackage{textcomp}
\usepackage{xcolor}
\usepackage{mathtools}
\usepackage{tabularx}
\usepackage{makecell}
\usepackage{dsfont}
\usepackage[absolute,overlay]{textpos}

\usepackage{hyperref}
\hypersetup{
    colorlinks=true,
    linkcolor=orange,
    filecolor=magenta,      
    urlcolor=orange,
    citecolor=orange,
}

\title{\LARGE \bf
Cooperative Autonomous Vehicles that Sympathize with Human Drivers
}

\author{Behrad Toghi$^{1}$, Rodolfo Valiente$^{1}$, Dorsa Sadigh$^{2}$, Ramtin Pedarsani$^{3}$, Yaser P. Fallah$^{1}$
\thanks{*This material is based upon work supported by the National Science Foundation under Grant No. CNS-1932037.}
\thanks{$^{1}$Connected \& Autonomous Vehicle Research Lab (CAVREL), University of Central Florida, Orlando, FL, USA. \tt\small {toghi@knights.ucf.edu}}%
\thanks{$^{2}$ Intelligent and Interactive Autonomous Systems Group (ILIAD), Stanford University, Stanford, CA, USA.}%
\thanks{$^{3}$ Department of Electrical and Computer Engineering, University of California, Santa Barbara, CA, USA.}}%
\begin{document}

\maketitle
\thispagestyle{empty}
\pagestyle{empty}
\begin{abstract}
Widespread adoption of autonomous vehicles will not become a reality until  solutions are developed that enable these intelligent agents to co-exist with humans. This includes safely and efficiently interacting with human-driven vehicles, especially in both conflictive and competitive scenarios. We build up on the prior work on socially-aware navigation and borrow the concept of social value orientation from psychology ---that formalizes how much importance a person allocates to the welfare of others--- in order to induce altruistic behavior in autonomous driving. In contrast with existing works that explicitly model the behavior of human drivers and rely on their expected response to create opportunities for cooperation, our Sympathetic Cooperative Driving (SymCoDrive) paradigm trains altruistic agents that realize safe and smooth traffic flow in competitive driving scenarios only from experiential learning and without any explicit coordination. We demonstrate a significant improvement in both safety and traffic-level metrics as a result of this altruistic behavior and importantly conclude that the level of altruism in agents requires proper tuning as agents that are too altruistic also lead to sub-optimal traffic flow. The code and supplementary material are available at: \href{https://symcodrive.toghi.net/}{\texttt{https://symcodrive.toghi.net/}}
\end{abstract}
%
\begin{textblock*}{20cm}(3.3cm,0.4cm) 
   \textit{Accepted in 2021 IEEE/RSJ International Conference on Intelligent Robots and Systems (IROS)}
\end{textblock*}
\section{Introduction}
\label{sec:introduction}
The next generation of transportation systems will be safer and more efficient with connected autonomous vehicles. Vehicle-to-vehicle (V2V) communication enables autonomous vehicles (AVs) to constitute a form of mass intelligence and overcome the limitations of a single agent planning in a decentralized fashion~\cite{palanisamy2020multi}. If all vehicles on the road were connected and autonomous, V2V could allow them to coordinate and handle complex driving scenarios that require selflessness, e.g., merging to and exiting a highway, and crossing intersections~\cite{toghi2018multiple, toghi2019analysis}. However, a road shared by AVs and human-driven vehicles (HVs) naturally becomes a competitive scene due to their different levels of maneuverability and reaction time. In contrast with the full-autonomy case, here the coordination between HVs and AVs is not as straightforward since AVs do not have an explicit means of harmonizing with humans and therefor require to locally account for the other HVs and AVs in their proximity.

\begin{figure}[t]
  \centering
  \includegraphics[width=.48\textwidth]{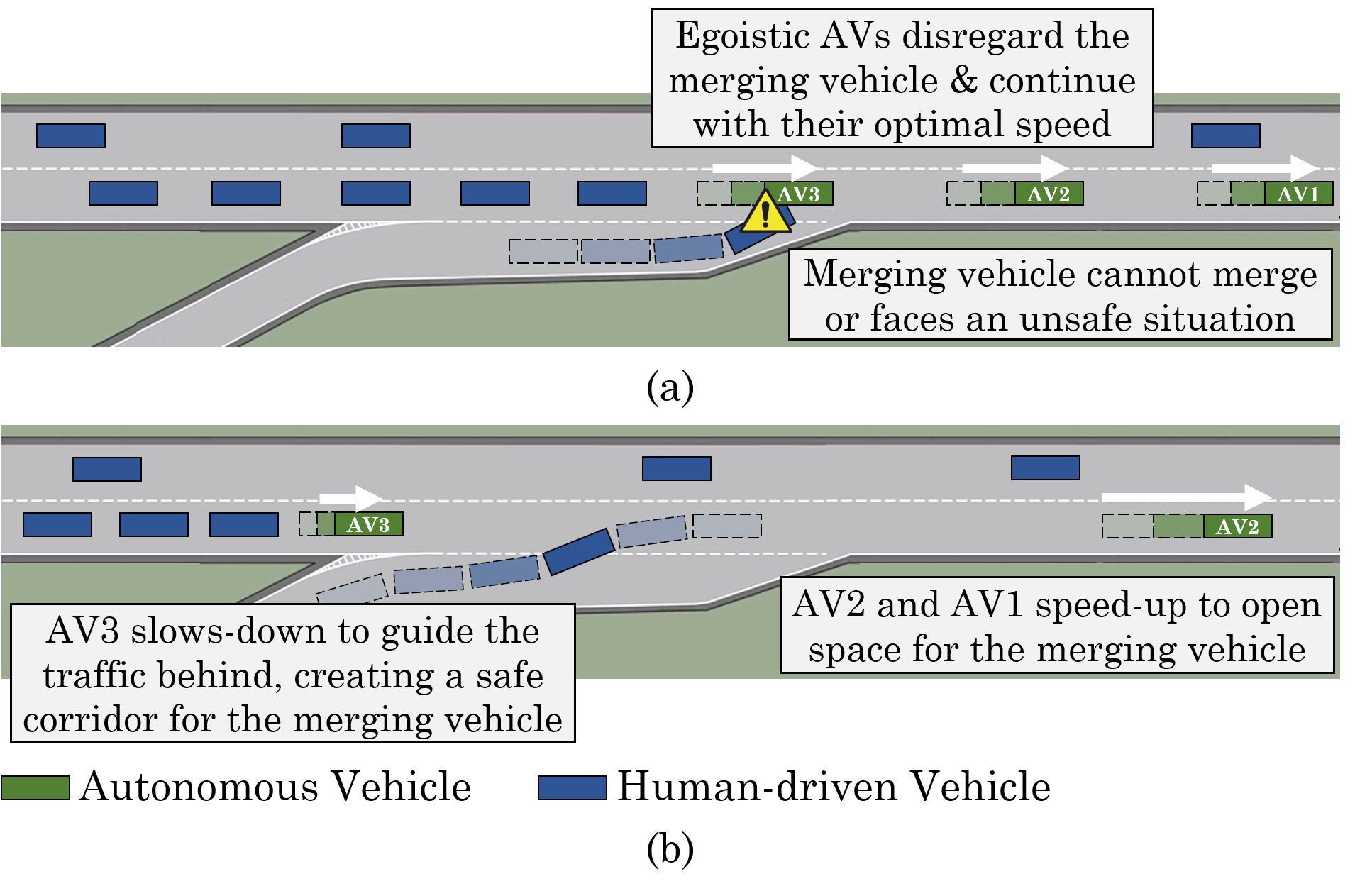}
  \caption{\small{Seamless and safe highway merging requires all AVs working together and accounting for the human-driven vehicles' utility. \emph{(top)} Egoistic AVs solely optimize for their own utility, \emph{(bottom)} Altruistic AVs compromise on their welfare to account for the human-driven vehicles.}}
  \label{fig:mainfigurecomparison}
\end{figure}

To further elaborate on this need, assume the merging scenario depicted in Figure~\ref{fig:mainfigurecomparison}. The merging vehicle, either HV or AV, faces a mixed group of AVs and HVs on the highway and needs them to slow-down to allow it to merge. If AVs act selfishly, it will be up to the HVs in the highway to allow for merging. Relying only on the human drivers can lead to sub-optimal or even unsafe situations due to their hard-to-predict and differing behaviors. In this particular example, assuming egoistic AVs, the merging vehicle will either get stuck in the merging ramp and not be able to merge or will wait for an HV and risk on cutting into the highway without knowing if the HV will slow-down or not. On the other hand, altruistic AVs can work together and guide the traffic on the highway, e.g., by slowing down the vehicles behind as AV3 does in Figure~\ref{fig:mainfigurecomparison}\hyperref[fig:mainfigurecomparison]{(b)}, in order to enable a seamless and safe merging. Such altruistic autonomous agents can create societally desirable outcomes in conflictive driving scenarios, without relying on or making assumptions about the behavior of human drivers.

%

Altruistic behavior of autonomous cars can be formalized by quantifying the willingness of each vehicle to incorporate the utility of others, whether an HV or an AV, into its local utility function. This notion is defined as \emph{social value orientation (SVO)}, which has recently been adopted from the psychology literature to robotics and artificial intelligence research~\cite{schwarting2019social}. SVO determines the degree to which an agent acts egoistic or altruistic in the presence of others. Figure~\ref{fig:mainfigurecomparison}\hyperref[fig:mainfigurecomparison]{(b)} demonstrates an example of altruistic behavior by AVs where they create a safe corridor for the merging HV and enable a seamless merging. In a mixed-autonomy scenario, agents either are homogeneous with the same SVO or can directly obtain each other's SVO (via V2V). However, the utility and SVO of an HV are unknowns, as they are subjective and inconstant and therefore cannot be communicated to the AVs. 

The existing social navigation works model a human driver's SVO either by predicting their behavior~\cite{alahi2016social} and avoiding conflicts with them or relying on the assumption that humans are naturally willing or can be incentivized to cooperate~\cite{sadigh2016planning}. By explicitly modeling human behavior, agents can exploit cooperation opportunities in order to achieve a social goal that favors both humans and autonomous agents. However, modeling human behaviors is often challenging due to time-varying changes in the model affected by fatigue, distraction, and stress as well as scalability of belief modeling techniques over other agent's behaviors, hence limiting the practicality of the above approach. Methods based on model-predictive control (MPC) generally require an engineered cost function and a centralized coordinator~\cite{guanetti2018control}. As such, they are not suitable for cooperative autonomous driving, where central coordination is not viable. On the other hand, data-driven solutions such as reinforcement learning are challenged in mixed-autonomy multi-agent systems, mainly due to the non-stationary environment in which agents are evolving concurrently.

Considering these shortcomings, the notion of altruism in AVs can be divided into \emph{cooperation} within autonomous agents and \emph{sympathy} among autonomous agents and human drivers. Dissociating the two components helps us to separately probe their influence on achieving a social goal. Our key insight is that defining a social utility function can induce altruism in decentralized autonomous agents and incentivize them to cooperate with each other and to sympathize with human drivers with no explicit coordination or information about the humans' SVO. The core differentiating idea that we rely on is that AVs trained to reach an optimal solution for all vehicles, learn to implicitly model the decision-making process of humans only from experience.
We study the behavior of altruistic AVs in scenarios that would turn into safety threats if either of sympathy and cooperation components is absent. In other words, we perform our experiments in scenarios with a similar nature to the one depicted in Figure~\ref{fig:mainfigurecomparison} that essentially require all agents to work together and success cannot be achieved by any of them individually.
Our main contributions are as follows:
\begin{itemize}
  \item We propose a data-driven framework, Sympathetic Cooperative Driving (SymCoDrive), that incorporates a decentralized reward structure to model cooperation and sympathy and employ a 3D convolutional deep reinforcement learning (DRL) architecture to capture the temporal information in driving data,
  \item We demonstrate how tuning the level of altruism in AVs leads to different emerging behaviors and affects the traffic flow and driving safety,
  \item We experiment with a highway merging scenario and demonstrate that our approach results in improved driving safety and societally desirable behaviors compared to egoistic autonomous agents.
\end{itemize}
\section{Related Work}
\label{sec:relatedworks}
\noindent \textbf{Multi-agent Reinforcement Learning. } 
A major challenge for multi-agent reinforcement learning (MARL) is the inherent non-stationarity of the environment. Foerster et al. suggest a novel learning rule to address this issue~\cite{foerster2017learning}. Additionally, the idea of decorrelating training samples by drawing them from an experience replay buffer becomes obsolete and a multi-agent derivation of importance sampling can be employed to remove the outdated samples from the replay buffer~\cite{foerster2017stabilising}. Xie et al. have also attempted to mitigate this problem by using latent representations of partner strategies to enable a more scalable MARL and partner modeling~\cite{xie2020learning}. 

The \textit{counterfactual multi-agent (COMA)} algorithm proposed by Foerster et al. uses a centralized critic and decentralized actors to tackle the problem of credit assignment in multi-agent environments~\cite{foerster2018counterfactual}. In the case of centralized control, deep Q-networks with full observability over the environment can be used to control the joint-actions of a group of agents~\cite{egorov2016multi}. Within the context of mixed-autonomy, the existing literature focuses on solving cooperative and competitive problems by making assumptions on the nature of interactions between autonomous agents (or autonomous agents and humans)~\cite{omidshafiei2017deep}. Contrary to these works, we assume partial observability and a decentralized reward function and aim to train sympathetic cooperative autonomous agents with no assumption on humans' behavior.

\smallskip
\noindent \textbf{Autonomous Driving in Mixed-autonomy. }
Driving styles of humans can be learned from demonstration through inverse RL or employing statistical models~\cite{sadigh2016planning, 8690965, 8690570, shah2019real}. Modeling human driver behavior assists autonomous vehicles to identify potentials for creating cooperation and interaction opportunities with humans in order to realize safe and efficient navigation~\cite{toghi2020maneuver}. Moreover, human drivers are able to intuitively anticipate next actions of neighboring vehicles through observing slight changes in their trajectories and leverage the prediction to move proactively if required. Inspired by this fact, Sadigh et al. reveal how autonomous vehicles can exploit this farsighted behavior of humans to shape and affect their actions~\cite{sadigh2016planning}. On a macro-traffic level, prior works have demonstrated emerging human behaviors within mixed-autonomy scenarios and studied how these patterns can be utilized to control and stabilize the traffic flow~\cite{wu2018stabilizing, lazar2019learning}. Closely related to our topic, recent works in social robot navigation have shown the potential for collaborative planning and interaction with humans as well~\cite{pokle2019deep, chen2017socially, alahi2016social}.
\vspace{-0.3cm}
\section{Preliminaries}
\label{sec:background}
\noindent \textbf{Partially Observable Stochastic Games (POSG). } We formulate the problem of multi-vehicle interaction using a stochastic game defined by the tuple $\mathcal{M}_\text{G} \coloneqq (\mathcal{I}, \mathcal{S}, [ \mathcal{A}_i ], [ \mathcal{O}_i ], P, [ r_i ])$ for $i=1,\dots, N$, in which $\mathcal{I}$ is a finite set of agents and $\mathcal{S}$ represents the state-space including all possible formations that the $N$ agents can adopt. At a given time the agent receives a local observation $o_i: \mathcal{S} \rightarrow \mathcal{O}_i$ and takes an action within the action-space $a_i \in \mathcal{A}_i$ based on a stochastic policy $\pi_i: \mathcal{O}_i \times \mathcal{A}_i \rightarrow [0, 1]$. Consequently, the agent transits to a new state $s'_i$ which is determined based on the state transition function $\Pr(s'|s, a): \mathcal{S} \times \mathcal{A}_1 \times ... \times \mathcal{A}_N \rightarrow \mathcal{S} $ and receives a reward $r_i: \mathcal{S} \times \mathcal{A}_i \rightarrow \mathbb{R}$. The goal is to derive an optimal policy $\pi^*$ that maximizes the discounted sum of future rewards over an infinite time horizon.

In a partially-observable stochastic game (POSG), the state transition and reward functions are usually not known and an agent only has access to a local observation which is correlated with the state. Employing multi-agent reinforcement learning, independent MARL agents can work together to overcome the physical limitations of a single agent and outperform them~\cite{tan1993multi}. In a multi-vehicle problem, controlling vehicles by a centralized MARL controller that has full observability over the environment and assigns a centralized joint reward ($\forall i,j: r_i \equiv r_j$) to all vehicles is rather straightforward. However, such assumptions are not feasible in real-world autonomous driving applications and we rather focus on the decentralized case where vehicles have partial observability and are not aware of each other's actions. Coordination among agents in such settings is expected to arise from the decentralized reward function that we introduce which uses the local observations to estimate the utility of other vehicles.

\smallskip
\noindent \textbf{Deep Q-networks (DQN). }
Q-learning, which has been widely applied in reinforcement learning problems with large state-spaces, defines a state-value function $Q^\pi(s,a) \coloneqq \mathbb{E} [\sum_{i=1}^\infty \gamma^i r(s_i, \pi (s_i)) |s_0=s, a_0=a]$ to derive the optimal policy $\pi^*(s) = \arg\max_a Q^* (s,a)$ where $\gamma \in [0,1)$ is a discount factor. DQN~\cite{mnih2013playing} uses a neural network with weights $\textbf{w}$ to estimate the state-action value function by performing mini-batch gradient descent steps as $\textbf{w}_{i+1} = \textbf{w}_i + \alpha_i \hat{\nabla}_\textbf{w} \mathcal{L}(\textbf{w}_i)$, where the loss function is defined as
\begin{equation}
\label{equ:Qlearning}
\mathcal{L}(\textbf{w}_i) = \mathbb{E}[(r+\gamma \: \underset{a}{\max} \: Q^*(s',a';\textbf{w}^\circ) - Q^*(s,a;\textbf{w})^2]
\end{equation}
and the $\hat{\nabla}_\textbf{w}$ operator is an estimate of the gradient at $\textbf{w}_i$ and $\textbf{w}^\circ$ is the target network's weights which get updated periodically in training. Sets of $(s, a, r, s')$ are randomly drawn from an \textit{experience replay buffer} to de-correlate the training samples in Equation~\eqref{equ:Qlearning}. This mechanism becomes problematic when agents' policies evolve during the training.
\vspace{-0.2cm}
\section{Sympathetic Cooperative Driving}
\label{sec:sycodrive}
\noindent \textbf{Highway Merging Scenario. }
Our base scenario is a highway merging ramp where a merging vehicle (either HV or AV) attempts to join a mixed platoon of HVs and AVs, as illustrated in Figure~\ref{fig:mainfigurecomparison}. We specifically choose this scenario due to its inherent competitive nature, since the local utility of the merging vehicle is conflictive with that of the cruising vehicles. We ensure that only one AV yielding to the merging vehicle will not make the merge possible and for it to happen, essentially all AVs require to work together. In Figure~\ref{fig:mainfigurecomparison}\hyperref[fig:mainfigurecomparison]{(b)}, AV3 must slow down and guide the vehicles in behind, which perhaps are not able to see the merging vehicle, while AV2 and AV1 speed-up to open space for the merging vehicle. If any of the vehicles do not cooperate or act selfishly, traffic safety and efficiency will be compromised.

\vspace{-0.1cm}
\smallskip
\noindent \textbf{Formalism. }
Consider a road section as shown in Figure~\ref{fig:mainfigurecomparison} with a set of autonomous vehicles $\mathcal{I}$, a set of human-driven vehicles $\mathcal{V}$, and a \emph{mission vehicle}, $M \in \mathcal{I} \cup \mathcal{V}$ that can be either AV or HV and is attempting to merge into the highway. HVs normally have a limited perception range restricted by occlusion and obstacles. In the case of AVs, although we assume no explicit coordination and no information about the actions of the others, autonomous agents are connected through V2V communication which allows them to share their situational awareness. Leveraging this extended situational awareness, agents can broaden their range of perception and overcome occlusion and line-of-sight visibility limitations. Therefore, while each AV has a unique partial observation of the environment, they can see all vehicles within their extended perception range, i.e., they can see a subset of AVs $\widetilde{\mathcal{I}} \subset \mathcal{I}$, and a subset of HVs $\widetilde{\mathcal{V}} \subset \mathcal{V}$.

In order to model a mixed-autonomy scenario, we deploy a mixed group of HVs and AVs to cruise on a highway and target to maximize their speed while maintaining safety. The contrast between humans and autonomous agents is that humans are solely concerned about their own safety while the altruistic autonomous agents attempt to optimize for the safety and efficiency of the group. Social value orientation gauges the level of altruism in an agent's behavior. In order to systematically study the interaction between agents and humans, we decouple the notion of \textit{sympathy} and \textit{cooperation} in SVO. Specifically, we consider the altruistic behavior of an agent with humans as sympathy and refer to the altruistic behavior among agents themselves as cooperation. One rationale behind this definition is the fact that the two are different in nature as the sympathetic behavior can be one-sided when humans are not necessarily willing to help the agents. Cooperation, however, is a symmetric quality since the same policy is deployed in all AVs and as we will see in our experiments, social goal of the group can be achieved regardless of the humans' willingness to cooperate.

\smallskip
\noindent \textbf{Decentralized Reward Structure. }
The local reward received by agent $I_i \in \mathcal{I}$ can be decomposed to
\begin{equation} \label{equ:reward1}
\begin{aligned}
R_i(s_i, a_i) ={} & R^E+R^C+R^S
\\={} & \lambda^E r^E_i(s_i, a_i) + \\
& \lambda^C \sum_j r^{C}_{i, j} (s_i, a_i) + \lambda^S \sum_k r^{S}_{i, k} (s_i, a_i)
\end{aligned}
\end{equation}
in which $j \in \widetilde{\mathcal{I}} \setminus \{I_i\}$, $k \in (\widetilde{\mathcal{V}} \cup \{M\}) \setminus (\mathcal{I} \cap \{M\})$. The level of altruism or egoism can be tuned by $\lambda^E$, $\lambda^C$, and $\lambda^S$ coefficients. The $r^E_i$ component in Equation~\eqref{equ:reward1} denotes the local driving performance reward derived from metrics such as distance traveled, average speed, and a negative cost for changes in acceleration to promote a smooth and efficient movement by the vehicle. The cooperative reward term, $r^{C}_{i, j}$ accounts for the utility of the observer agent's allies, i.e., other AVs in the perception range except for $I_i$. It is important to note that $I_i$ only requires the V2V information to compute $R^{C}$ and not any explicit coordination or knowledge of the actions of the other agents. The sympathetic reward term, $r^{S}_{i, k}$ is defined as
\begin{equation} \label{equ:symreward}
r^{S}_{i, k} = r^M_k + \sum_k \frac{1}{\eta d_{i,k}^\psi}u_k,
\end{equation}
where $u_k$ denotes an HV's utility, e.g., its speed, $d_{i,k}$ is the distance between the observer autonomous agent and the HV, and $\eta$ and $\psi$ are dimensionless coefficients. Moreover, the sparse scenario-specific \textit{mission reward} term $r^M_k$ in the case of our driving scenario is representing the success or failure of the merging maneuver, formally
\begin{equation}
\label{equ:missionreward}
r^M_k = 
\begin{cases}
1,              & \text{if $V_k$ is the mission vehicle and has merged} \\
0,                  & \text{o.w.}
\end{cases}
\end{equation}

During training, each agent optimizes for this decentralized reward function using Deep RL and learns to drive on the highway and work with its allies to create societally desirable formations that benefits both AVs and HVs.

\smallskip
\noindent \textbf{State-space and Action-space. }
The robot navigation problem can be viewed from multiple levels of abstraction: from the low-level continuous control problem to the higher level meta-action planning. Our purpose in this work is to study the inter-agent and agent-human interactions as well as the behavioral aspects of mixed-autonomy driving. Thus, we choose a more abstract level and define the action-space as a set of discrete meta-actions $a_i \in \mathbb{R}^n$.

\begin{figure}[t]
  \centering
  \includegraphics[width=.48\textwidth]{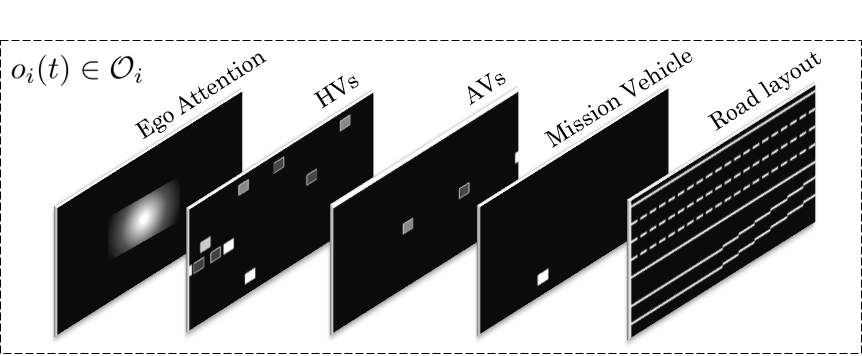}
\caption{\small{Multi-channel VelocityMap state representation embeds the speed of the vehicle in pixel values.}}
  \label{fig:heatmap}
\end{figure}

We experiment with two different local state representations to find the most suitable one for our problem. The \textit{multi-channel VelocityMap} representation separates AVs and HVs into two channels and embeds their relative speed in the pixel values. Figure~\ref{fig:heatmap} illustrates an example of this multi-channel representation. A clipped logarithmic function is used to map the relative speed of the vehicles into pixel values as it showed a better performance compared to the linear mapping, i.e.,
\begin{equation} \label{equ:logmapping}
Z_j = 1 - \beta \log (\alpha |v_j^{(l)}|) \mathds{1}(|v_j^{(l)}|-v_0) 
\end{equation}
where $Z_j$ is the pixel value of the $j$th vehicle in the state representation, $v^{(l)}$ is its relative Frenet longitudinal speed from the $k$th vehicle's point-of-view, i.e., $\dot{l_j}-\dot{l_k}$, $v_0$ is speed threshold, $\alpha$ and $\beta$ are dimensionless coefficients, and $\mathds{1}(.)$ is the Heaviside step function. Such non-linear mapping gives more importance to neighboring vehicles with smaller $|v^{(l)}|$ and almost disregards the ones that are moving either much faster or much slower than the ego. We add three more channels that embed 1) the road layout, 2) an attention map to emphasize on the location of the ego, and 3) the mission vehicle.

The other candidate is an \textit{occupancy grid} representation that directly embeds the information as elements of a 3-dimensional tensor $o_i \in \mathcal{O}_i$. Theoretically, this representation is very similar to the previous VelocityMap and what contrasts them is that the occupancy grid removes the shapes and visual features such as edges and corners and directly feeds the network with sparse numbers. More specifically, consider a tensor of size $W \times H \times F$, in which the $n$th channel is a $W \times H$ matrix defined as

\begin{equation}
\label{equ:occupancygrid}
o_{(n, \, ,\, )} \in \mathbb{R}^2 = \begin{cases}
f(n),              & \text{if} \, \, \,f(1) = 1\\ 
0,                  & \text{o.w.}
\end{cases}
\end{equation}
where $f=[p, l, d, v^{(l)}, v^{(d)}, \sin \delta, \cos \delta]$ is the feature set, $p$ is a binary variable showing the presence of a vehicle, $l$ and $d$ are relative Frenet coordinates, $v^{(l)}$ and $v^{(d)}$ are relative Frenet speeds, and $\delta$ is the yaw angle measured with respect to a global reference.

\begin{figure}[t]
  \centering
  \includegraphics[width=.48\textwidth]{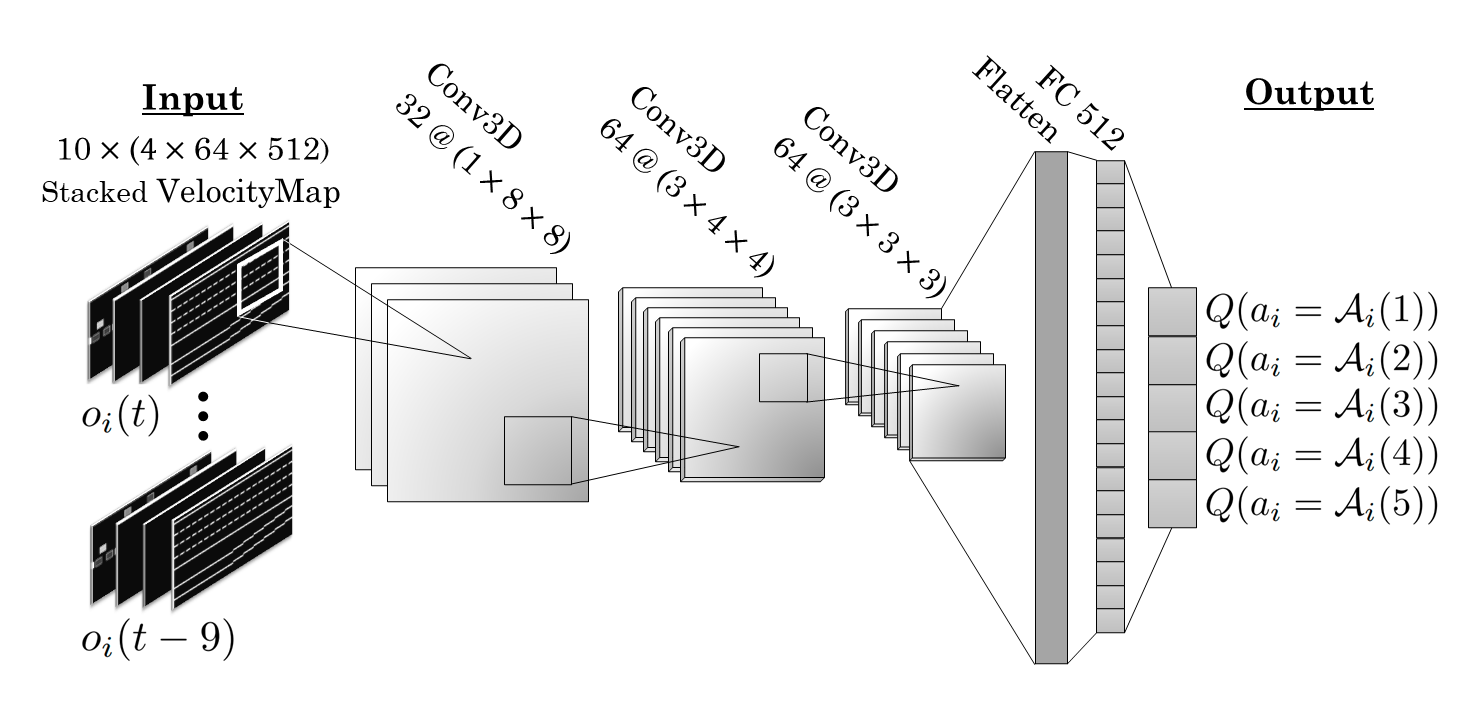}
  \caption{\small{Our deep Q-network with 3D Convolutional Architecture.}}
  \label{fig:architecture}
\end{figure}

\smallskip
\noindent \textbf{Training with Deep MARL. }
We experiment with 3 existing architectures proposed in the literature by Toghi et al., Mnih et al., and Egorov et al. as function approximators for our Q-learning problem~\cite{toghi2021social, mnih2013playing, egorov2016multi}. Additionally, we implemented a 3D convolutional network that captures the temporal dependencies in a training episode as shown in Figure~\ref{fig:architecture}. The input to our network is a stack of 10 VelocityMap observations, i.e., a $10 \times (4 \times 512\times 64)$ tensor, which capture the last 10 time-steps in the episode. The performance of the architectures is compared in Section~\ref{sec:deepnetworks}.

\begin{figure}[t]
  \centering
  \includegraphics[width=.48\textwidth]{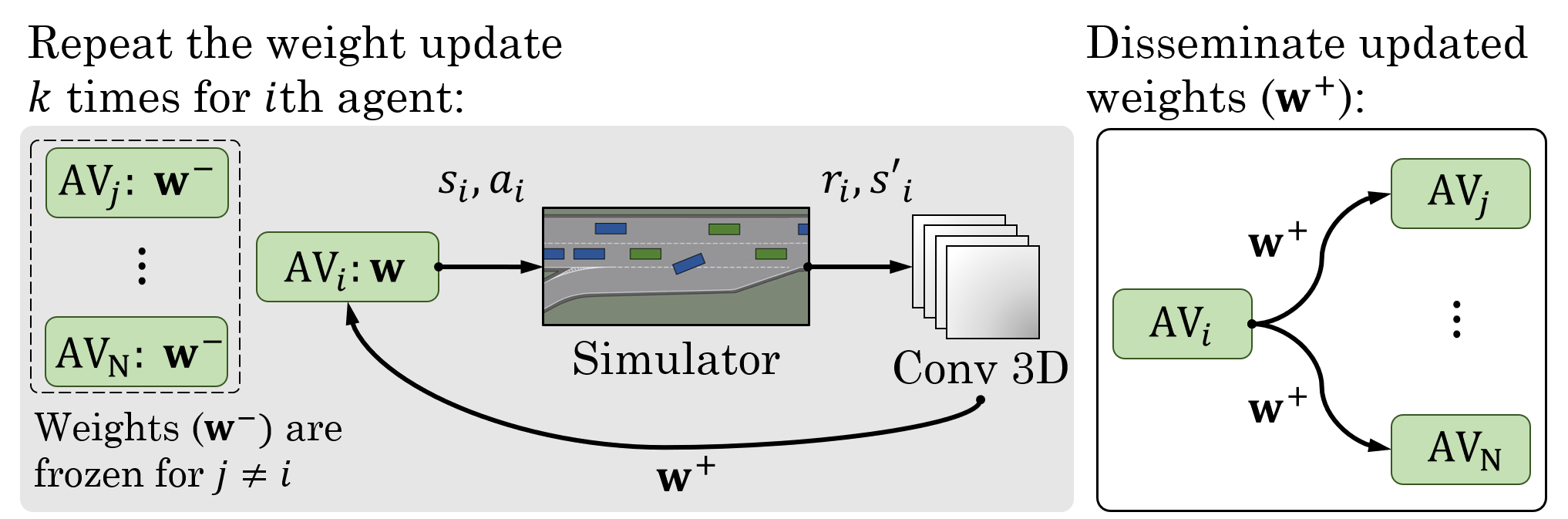}
  \caption{\small{The multi-agent training and policy dissemination process.}}
  \label{fig:coordinatedPOSG}
\end{figure}

We train a single neural network offline and deploy the learned policy into all agents for distributed independent execution in real-time. In order to cope with the non-stationarity issue in MARL, agents are trained in a semi-sequential manner, as illustrated in Figure~\ref{fig:coordinatedPOSG}. Each agent is trained separately for $k$ episodes while the policies of its allies, $\textbf{w}^-$, are frozen. The new policy, $\textbf{w}^+$, is then disseminated to all agents to update their neural networks. Additionally, inspired by~\cite{schaul2015prioritized}, we employ a novel experience replay mechanism to compensate for our highly skewed training data. A training episode can be semantically divided into two sections, cruising on a straight highway and highway merging. The ratio of the latter to the former in the experience replay buffer is a small number since the latter occurs in only a short time period of each episode. Consequently, uniformly sampling from the experience replay buffer leads to too few training samples relating to highway merging. Instead, we set the probability of a sample being drawn from the buffer proportional to its last resulted reward and its spatial distance with the merging point on the road. Balancing skewed training datasets is a common practice in computer vision and machine learning and appeared to be beneficial in our MARL problem as well.
%

\vspace{-0.15cm}
\section{Experiments}
\label{sec:experiments}
\vspace{-0.15cm}
\subsection{Driving Simulator Setup}
\vspace{-0.1cm}
We customize an OpenAI Gym environment~\cite{leurent2019approximate} to simulate the highway driving and merging scenarios. In the framework of our simulator, a Kinematic Bicycle Model describes the motion of the vehicles and a closed-loop proportional–integral–derivative (PID) controller is employed for translating the meta-actions to low-level steering and acceleration control signals. Particularly, we choose a set of $n=5$ abstract actions as
$a_i \in \mathcal{A}_i = [\texttt{Lane Left}$, $\texttt{Idle}$, $\texttt{Lane Right}$, $\texttt{Accelerate}$, $\texttt{Decelerate}]^T$.
As a common practice in the autonomous driving space, we express road segments and vehicles' motion in the Frenet-Serret coordinate frame which helps us to take the road curvature out of our equations and break-down the control problem to lateral and longitudinal components. In our simulated environment, the behavior of HVs is governed by lateral and longitudinal driver models proposed by Treiber et al. and Kesting et al~\cite{treiber2000congested, kesting2007general}.

In order to ensure the generalization capability of our learned policies, we draw the initial position of all vehicles from a clipped Gaussian distribution with mean and variance tuned to ensure that the initialized simulations fall into our desired merging scenario configuration. We further randomize the speed and initial position of the vehicles during the testing phase to probe the agents' ability to handle unseen and more challenging cases.

\vspace{-0.2cm}
\subsection{Computational Details}
\vspace{-0.1cm}
 A single training iteration in the PyTorch implementation of SymCoDrive takes about 440ms using a NVIDIA Tesla V100 GPU and a Xeon 6126 CPU @ 2.60GHz. We have repeated the training process multiple times to ensure all runs converge to similar emerging behaviors and policy. Training the Conv3D network for 15,000 episodes took approximately 33 hours on our hardware. The policy execution frequency is set to 1Hz and an online query of the network in the testing phase takes approximately 10ms. We spent $\sim$4,650 GPU-hours to tune the neural networks and reward coefficients for the purpose of our experiments.
\begin{figure*}[t]
  \centering
  \includegraphics[width=.98\textwidth]{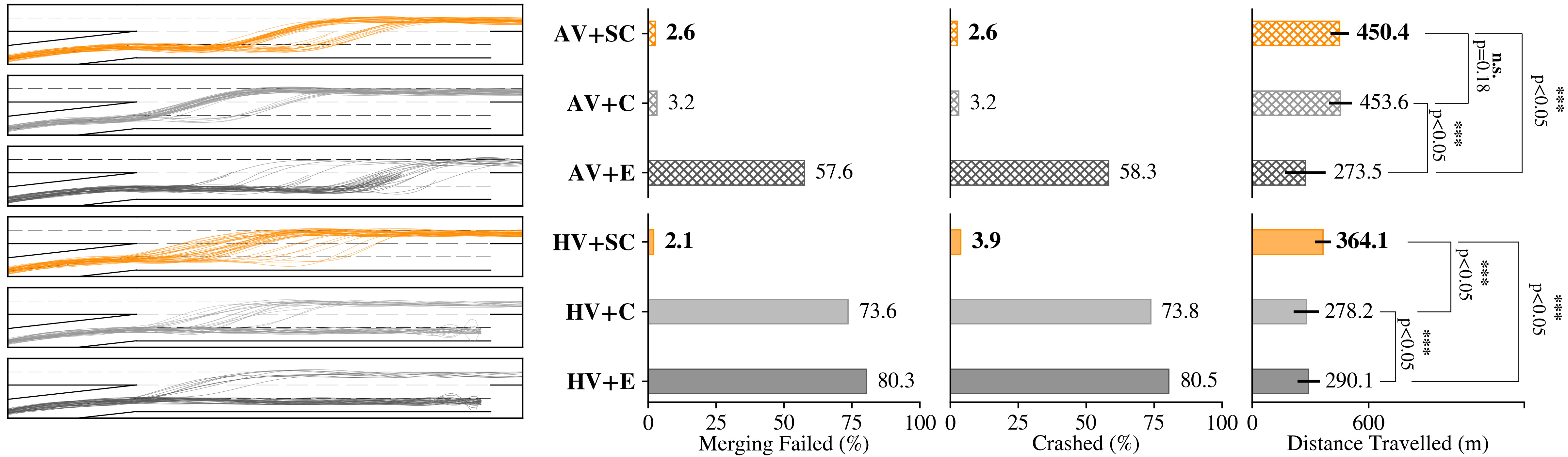}
  \caption{\small{Comparison between egoistic, cooperative-only, and sympathetic cooperative autonomous agents and how they interact with an autonomous (\textit{top}) or human-driven (\textit{bottom}) mission vehicle. A set of sampled mission vehicle's trajectories are illustrated on the left-side, relating to each of the 6 experiment setups defined in Section \ref{indepvars}. 
%
}}
  \label{fig:coopvsnoncoop}
\end{figure*}

\subsection{Independent Variables}
\label{indepvars}
We conducted a set of experiments to study how \emph{sympathy} and \emph{cooperation} components of the reward function impact the behavior of autonomous agents and the overall safety\slash efficiency metrics. We compare the case in which the mission vehicle---merging vehicle in the example in Fig.~\ref{fig:mainfigurecomparison}---is \emph{autonomous} to its dual scenario with a \emph{human-driven} mission vehicle. We define 2x4 settings, in which the mission vehicle is either an AV or HV, and the other autonomous agents follow an egoistic, cooperative-only, sympathetic-only, or sympathetic cooperative objectives: 
\begin{itemize}
    \item \textbf{HV+E. } The mission vehicle is \emph{human-driven} and autonomous agents act \emph{egoistically},
    \item \textbf{HV+C. } The mission vehicle is \emph{human-driven} and autonomous agents only have a \emph{cooperation} component ($R^C$) in their reward,
    \item \textbf{HV+S. } The mission vehicle is \emph{human-driven} and autonomous agents only have the \emph{sympathy} ($R^S$) element,
    \item \textbf{HV+SC. } The mission vehicle is \emph{human-driven} and autonomous agents have both \emph{sympathy} ($R^S$) and \emph{cooperation} ($R^C$) components in their reward,
    \item \textbf{AV+E/C/S/SC.} Similar to the cases above with the difference of mission vehicle being \emph{autonomous}.
\end{itemize}
\vspace{-0.1cm}
\subsection{Dependent Measures}
Performance of our experiments can be gauged in terms of efficiency and safety. The average distance traveled by each vehicle within the duration of a simulation episode is a traffic-level measure for efficiency. The percentage of the episodes that experienced a crash indicates the safety of the policy. Counting the number of scenarios with no crashes and successful missions (merging to the highway) gives us an idea about our solution's overall efficacy.
\vspace{-0.2cm}
\subsection{Hypotheses}
We examine three key hypotheses:
\begin{itemize}
    \item \textbf{H1.} \emph{In the absence of both cooperation and sympathy, a HV will not be able to safely merge into the highway. Thus, we anticipate to witness a better performance in \textbf{HV+SC} compared to \textbf{HV+C} and \textbf{HV+E}.}
    \item \textbf{H2.} \emph{An autonomous mission vehicle only requires altruism from its allies to successfully merge. We do not expect to see a significant difference between \textbf{AV+SC} and \textbf{AV+C} scenarios; however, we hypothesize that they both will outperform \textbf{AV+E}.}
    \item \textbf{H3.} \emph{Tuning the level of altruism in agents leads to different emerging behaviors that contrast in their impact on efficiency and safety. Increasing the level of altruism can become self-defeating as it jeopardizes the agent's ability to learn the basic driving skills.}
\end{itemize}
\vspace{-0.3cm}
\subsection{Results}
\vspace{-0.2cm}
We train SymCoDrive agents for 15,000 episodes in randomly initialized scenarios with a small standard deviation and average the performance metrics over 3,000 test episodes with 4x larger initialization range to ensure that our agents are not over-fitting on the seen training episodes.
\subsubsection{Cooperation \& Sympathy}
\begin{figure}[b]
  \centering
  \includegraphics[width=.44\textwidth]{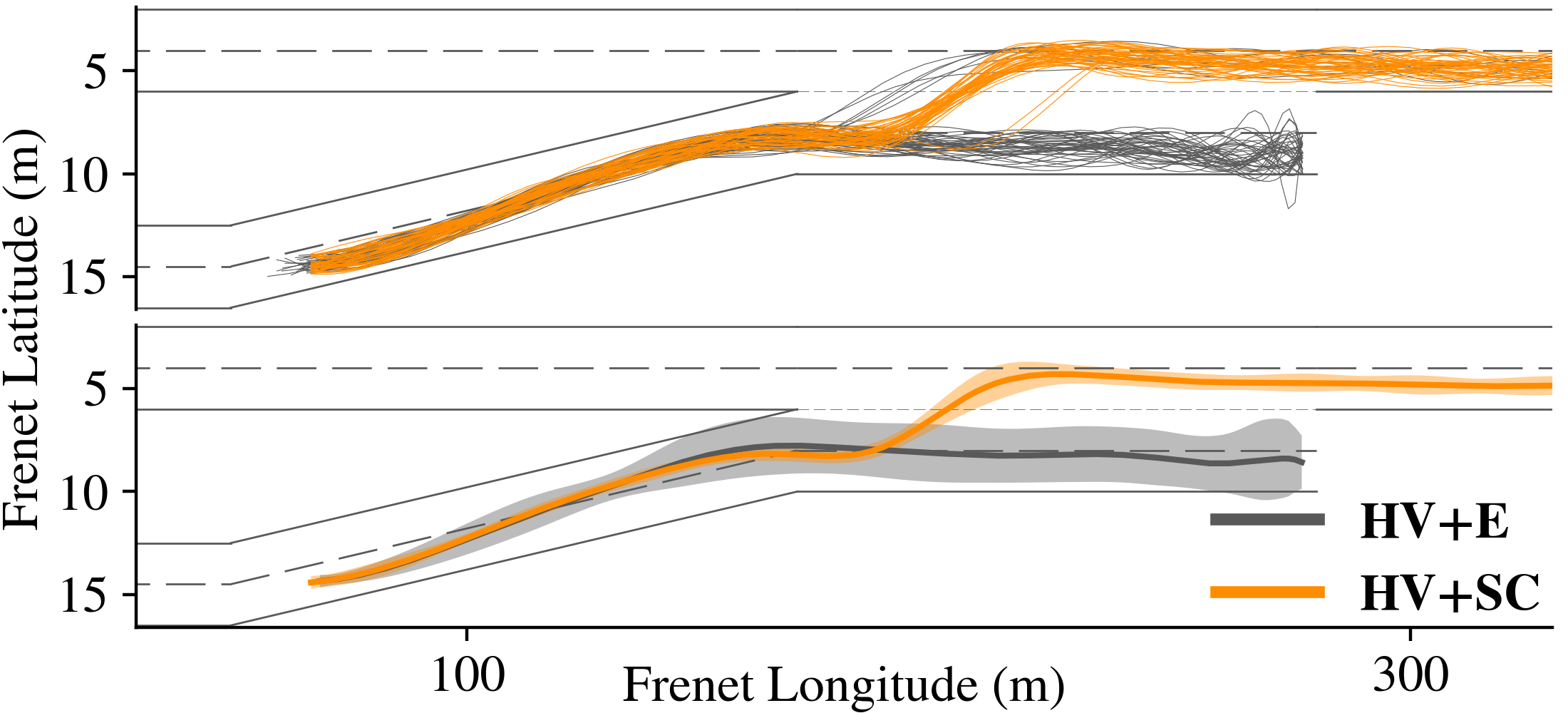}
  \caption{\small{A set of sample trajectories of the merging vehicle shows mostly successful merging attempts in \textbf{HV+SC}, compared to the failed attempts in \textbf{HV+E}.}}
  \label{fig:traj}
\end{figure}

To examine our hypothesis \textbf{H1}, we focus on scenarios with a human-driven mission vehicle, i.e., \textbf{HV+E}, \textbf{HV+C}, and \textbf{HV+SC}. The bottom row in Figure~\ref{fig:coopvsnoncoop} illustrates our observations for these scenarios. It is evident that agents that integrate cooperation and sympathy elements (\textbf{SC}) in their reward functions show superior performance compared to solely cooperative (\textbf{C}) or egoistic (\textbf{E}) agents. This insight is also reflected in the bar plots that measure the average distance traveled by vehicles on the bottom right-most side. As a result of fair and efficient traffic flow, vehicles in the \textbf{HV+SC} scenario clearly succeed to travel a longer distance whereas in the \textbf{HV+C} and \textbf{HV+E} scenarios failed merging attempts and possible crashes deteriorate the performance. The left-most column in Figure~\ref{fig:coopvsnoncoop} visualizes a set of sampled mission vehicle trajectories. It is clear that in the majority of episodes, cooperative sympathetic agents successfully merge to the highway while the other (\textbf{C}) and (\textbf{E}) agents fail in most of their attempts. Figure~\ref{fig:traj} provides further intuition on our discussion by comparing a set of mission vehicle's trajectories extracted from a \textbf{HV+E} scenario to the trajectories from the \textbf{HV+SC} scenario. Evidently, cooperative sympathetic agents enable successful merging while the other egoistic and solely-cooperative agents fail to do so, supporting our hypothesis \textbf{H1}.

It is imperative to repeat the experiments above for scenarios with an autonomous mission vehicle as one can argue that the failed missions and crashes in \textbf{HV+C} and \textbf{HV+E} are due to inadequacy of the driver model we have chosen for HVs. To precisely address this argument, \textbf{AV+E}, \textbf{AV+C}, and \textbf{AV+SC} scenarios are illustrated in the top row of Figure~\ref{fig:coopvsnoncoop}. First, a comparison between two scenarios with egoistic agents, i.e., \textbf{AV+E} and \textbf{HV+E}, unveils that an autonomous mission vehicle acts more creatively and explores different ways of merging to the highway, hence the more spread trajectory samples in \textbf{AV+E} compared to \textbf{HV+E}. Next, comparing the performance of an egoistic autonomous mission vehicle with a human-driven mission vehicle in terms of crashes and failed merges shows the autonomous agent is generally more capable to find a way to merge into the platoon of humans and egoistic agents. However, it still fails in more than half of its merging attempts. Figure~\ref{fig:coopvsnoncoop} verifies our hypothesis \textbf{H2} as we can observe that adding only a cooperation component to the agents, i.e., \textbf{AV+C} scenario, enables the mission vehicle to merge to the highway almost in all of its attempts. Adding the sympathy element in \textbf{AV+SC} slightly improves the safety as it incentivizes the agents to be aware of the humans that are not in the direct risk of collision with them.
We consider cooperation as an enabler for sympathy and did not conduct any experiment with sympathetic-only setting as its results can be inferred from a comparison between (\textbf{SC}) and (\textbf{C}).
\begin{figure}[t]
  \centering
  \includegraphics[width=.48\textwidth]{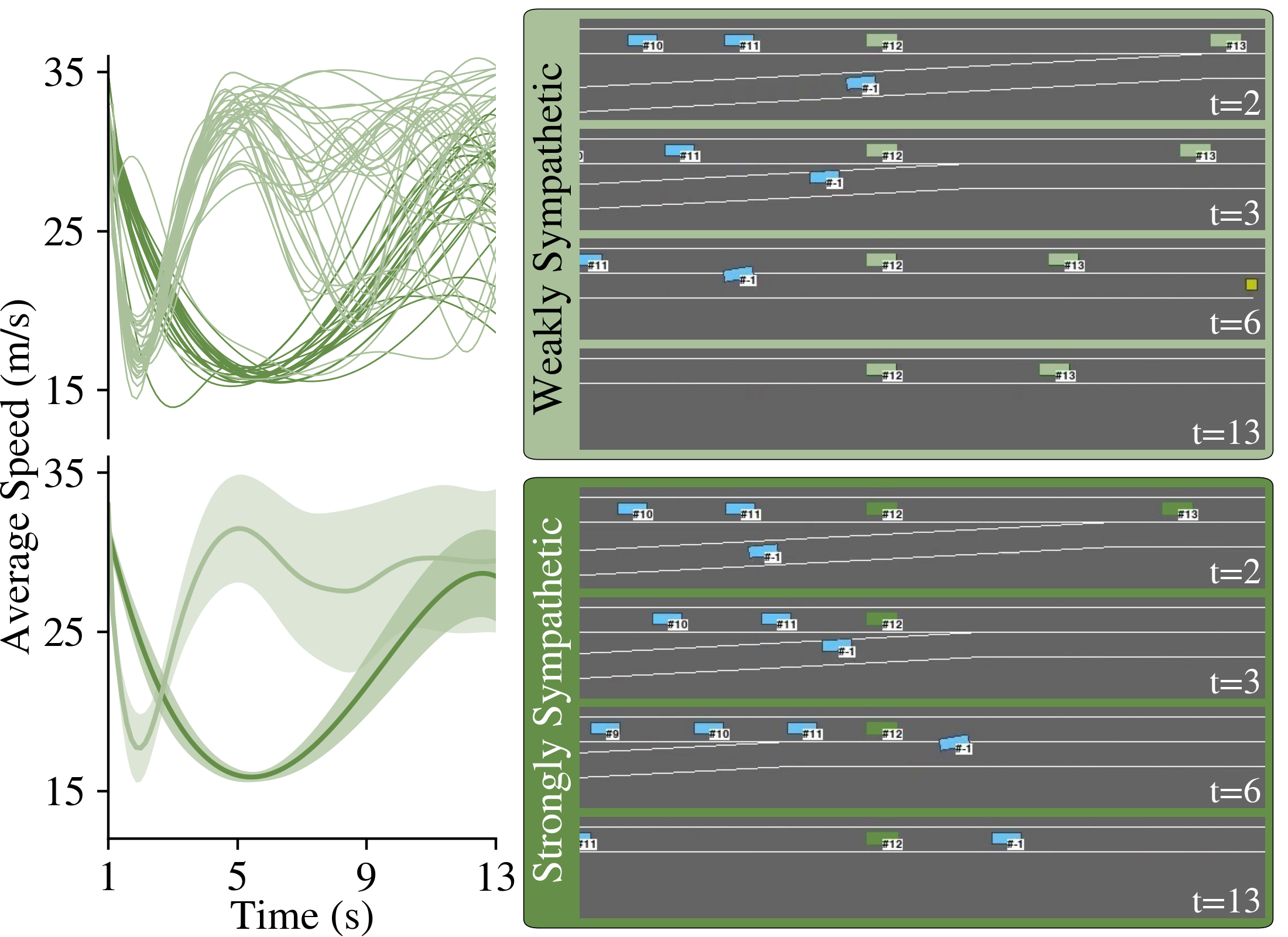}
  \caption{\small{Comparing weakly and strongly sympathetic autonomous agents: (\textit{left}) Speed profiles of the "guide AV" (consider AV3 in Figure~\ref{fig:mainfigurecomparison}\hyperref[fig:mainfigurecomparison]{(b)}) and (\textit{right}) Sample snapshots.}}
  \label{fig:speedprofiles_snapshot}
\end{figure}

\begin{figure}[b]
  \centering
  \includegraphics[width=.4\textwidth]{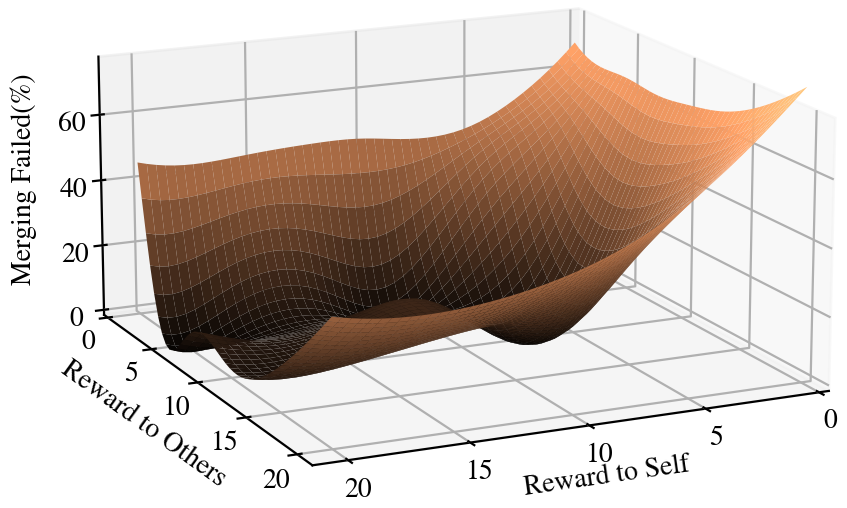}
  \caption{\small{Tuning SVO for autonomous agents reveals that an optimal point between \emph{caring about others} and \emph{being selfish} exists that eventually benefits all the vehicles in the group.}}
  \label{fig:svo}
\end{figure}

\subsubsection{Tuning Altruism \& Emerging Behaviors}
To investigate hypothesis \textbf{H3}, we train a set of agents and vary their reward coefficients, i.e., $\lambda_E$, $\lambda_C$, $\lambda_S$, to adjust their level of sympathy and cooperation. Revisiting our driving scenario depicted in Figure~\ref{fig:mainfigurecomparison}, we particularly witness two critical emerging behaviors in agents. Strongly sympathetic agents that are trained with a high ratio of $\lambda_S/(\lambda_C+\lambda_E)$, naturally prioritize the benefit of humans over their own. Figure~\ref{fig:speedprofiles_snapshot} shows a set of snapshots extracted from two scenarios with strongly sympathetic and weakly sympathetic agents. A strongly sympathetic agent (consider AV3 in Figure~\ref{fig:mainfigurecomparison}\hyperref[fig:mainfigurecomparison]{(b)}) slows down and blocks the group of vehicles behind it to ensure that the mission vehicle gets a safe pathway to merge. On the other hand, the weakly sympathetic agent initially brakes to slow down the group of the vehicles behind it and then prioritizes its own benefit, speeds up, and passes the mission vehicle. Although both behaviors enable the mission vehicle to successfully merge, the speed profiles of the agent in Figure~\ref{fig:speedprofiles_snapshot} depict how a strongly sympathetic agent compromises on its traveled distance (the area under the speed curve) to maximize the mission vehicle's safety. Motivated by this observation, we thoroughly studied the effect that tuning the reward coefficients in Equation~\eqref{equ:reward1} makes on the performance of SymCoDrive agents. As illustrated in Figure~\ref{fig:svo}, we empirically observe that an optimal point between \emph{caring about others} and \emph{being selfish} exists that eventually benefits all the vehicles in the group.

\vspace{-0.2cm}
\subsection{Deep Networks and Generalization}
\begin{figure}[t]
\centering
\includegraphics[width=0.48\textwidth]{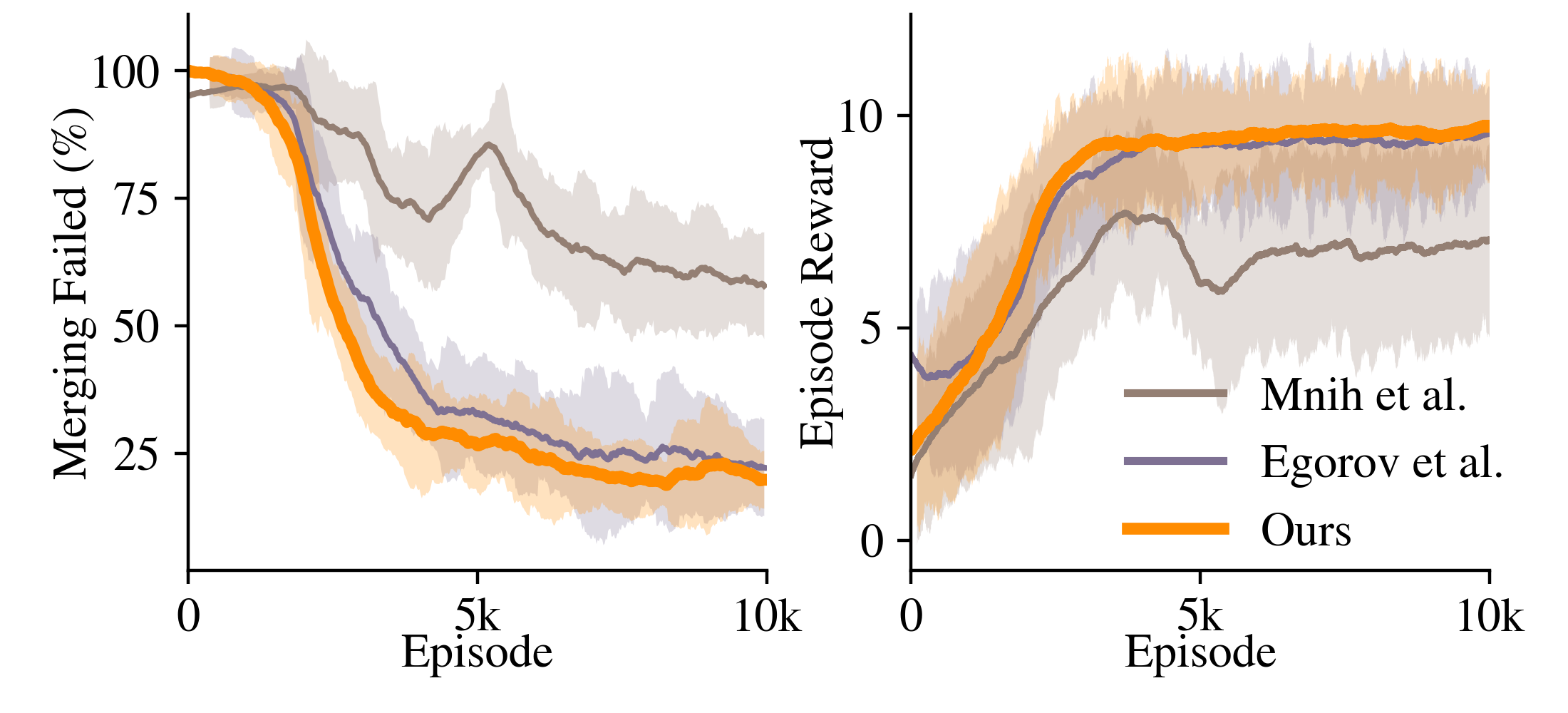}
\caption{\small{Training performance of the three benchmark network architectures.}}
\label{fig:trainingsummary}
\end{figure}

\vspace{-0.1cm}
\label{sec:deepnetworks}
We trained the network architectures introduced in Section~\ref{sec:sycodrive} and examined their ability in generalizing to test episodes with 4$\times$ wider range of initialization randomness, figure~\ref{fig:trainingsummary} shows the training performance of the networks. When tested in episodes with the same range of initialization randomness as training, all networks showed acceptable performance. However, their performance quickly depreciated when the range of randomness was increased and agents faced episodes different than what they had seen during the training, as noted in Table~\ref{table:comp}. While the other networks over-fitted on the training episodes, our Conv3D architecture significantly outperformed them in the more diverse test scenarios. We conclude that using VelocityMaps and our Conv3D architecture, agents learn to handle more complex unseen driving scenarios. Table~\ref{table: hyperparameters} lists the hyper-parameters we have used to train our Conv3D architecture.

%

\begin{table*}[t]
\caption{\small{Performance comparison of related architectures. Our Conv3D architecture outperformed the others as the level of randomness increases and agents face episodes different than what they had seen during the training.}}
\begin{center}
\begin{tabular}{c c c c|  c c c |  c c c  }
&
\multicolumn{3}{c}{Low Randomness} &
\multicolumn{3}{c}{Medium Randomness} &
\multicolumn{3}{c}{High Randomness}\\
Models &
C (\%)&
MF (\%) &
DT (m)&
C (\%)&
MF (\%) &
DT (m)&
C (\%)&
MF (\%) &
DT (m)\\
\hline
Toghi et al.~\cite{toghi2021social}& 6.2 &
\textbf{0} &
288 &
65.2 &
65.2 &
304 &
78.9 &
31.4 &
212 \\
Mnih et al.~\cite{mnih2013playing} &
9.6 &
7.2 &
\textbf{350} &
41.2 &
41.2 &
240 &
12.9 &
10.8 &
344 \\
Egorov et al. ~\cite{egorov2016multi} &
19.7 &
9.0 &
312 &
7.3 &
1.7 &
366 &
18.9 &
8.4 &
313 \\
\textcolor{black}{\textbf{Conv3D (Ours)}} & 
\textcolor{black}{\textbf{3.3}} & 
\textcolor{black}{0.2} & 
\textcolor{black}{334}& 
\textcolor{black}{\textbf{2.4}} & 
\textcolor{black}{\textbf{0.4}} & 
\textcolor{black}{\textbf{373}} & 
\textcolor{black}{\textbf{4.8}} & 
\textcolor{black}{\textbf{1.0}} & 
\textcolor{black}{\textbf{351}}\\
\hline
\hline
\end{tabular}
\end{center}
\raggedright\footnotesize{\hspace{0.4cm} \quad \quad \quad \quad \quad   C: \emph{Crashed}, MF: \emph{Merging Failed}, DT: \emph{Distance Travelled}}\\
\label{table:comp}
\end{table*}

The Occupancy Grid state-space representation, defined in Equation~\eqref{equ:occupancygrid} showed an inferior performance in all neural network architectures compared to the VelocityMap representation in our particular driving problem. We speculate that this is due to the fact that the Occupancy Grid representation does not benefit from the road layout and visual cues embedded in the VelocityMap state representation. All of our experiments discussed earlier are performed with VelocityMap representation, unless stated otherwise. After tuning the VelocityMaps, we concluded that integrating a hard ego-attention map in the state representation did not make a significant enhancement and decided to drop this channel, reducing the number of channels to~4. Instead, we aligned the center of VelocityMaps with regards to the ego such that 30\% of the observation frame reflects the range behind the ego and the rest shows the range in front. We noticed that this parameter plays an important role in training convergence and the resulted behaviors as it enables the agent to see the mission vehicle and other vehicles before they get to its close proximity.

\vspace{-0.3cm}
\section{Concluding Remarks}
\noindent \textbf{Summary. }
We tackle the problem of autonomous driving in mixed-autonomy environments where autonomous vehicles interact with vehicles driven by humans. We incorporate a cooperative sympathetic reward structure into our MARL framework and train agents that cooperate with each other, sympathize with human-driven vehicles, and consequently demonstrate superior performance in competitive driving scenarios, such as highway merging, compared to egoistically trained agents.

\smallskip
\noindent \textbf{Limitations and Future Work. }
Our current reward structure includes a hand-crafted marker that depends on the driving scenario, e.g., merging or exiting a highway. Given diverse driving episodes, this marker can also be learned from interaction data, cutting the need for a mission-specific reward term. We believe the merging scenario is representative of many common interaction scenarios we observe including other behaviors that require the two agents regulating their speeds and coordinating with each other such as exiting a highway.
We have only experimented with training and testing agents in the same scenario and have not cross-validated them across different scenarios. We hope to extend this work to other scenarios in the future. We believe, given a large enough training data, an agent is expected to learn the same altruistic behavior in general driving scenarios.
\begin{table}[t]
\caption{\small{List of hyper-parameters of our Conv3D Q-Network}}
\begin{center}
\begin{tabular}{c c | c c}
Hyper-param &
Value &
Hyper-param &
Value\\ 
\hline
\hline
Training iterations &
720,000 &
Initial exploration &
1.0 \\ 
Batch size &
32 &
Final exploration &
0.1 \\ 
Replay buffer size &
10,000 &
$\epsilon$ decay &
Linear \\ 
Learning rate &
0.0005 &
Optimizer &
ADAM \\ 
Target network update &
200 &
Discount factor $\gamma$ &
0.95 \\ 
\hline
\hline
\end{tabular}
\end{center}
\label{table: hyperparameters}
\end{table}

\vspace{-0.2cm}

\bibliographystyle{IEEEtran}
\bibliography{IEEEbibs} 
\vspace{-0.1cm}

\end{document}